\documentclass{article}

\usepackage{microtype}
\usepackage{graphicx}
\usepackage{subcaption}
\usepackage{booktabs} 
\usepackage{tabularx}
\usepackage{booktabs}
\usepackage{adjustbox} 
\usepackage{hyperref}
\usepackage{tabularx}
\usepackage{booktabs}
\usepackage{array}
\usepackage{makecell}
\usepackage{float}
\usepackage[preprint]{icml2026}

\usepackage{amsmath}
\usepackage{amssymb}
\usepackage{mathtools}
\usepackage{amsthm}
\usepackage{tabularx}

\usepackage[capitalize,noabbrev]{cleveref}

\theoremstyle{plain}

\theoremstyle{definition}

\theoremstyle{remark}

\usepackage[disable,textsize=tiny]{todonotes}

\icmltitlerunning{PC-MCMC-CIGP for Reaction Network Discovery}

\begin{document}

\twocolumn[
  \icmltitle{Synergizing Physically Constrained MCMC and Chemical-Informed Gaussian Processes for Reaction Network Discovery}

  \icmlsetsymbol{equal}{*}

  \begin{icmlauthorlist}
    \icmlauthor{Runzhe Liu}{equal,dut1,dut2}
    \icmlauthor{Zihao Wang}{equal,hkust}
    \icmlauthor{Wenbo Yang}{dut1,dut2}
    \icmlauthor{Shengyang Tao}{dut1,dut2}
  \end{icmlauthorlist}

  \icmlaffiliation{dut1}{State Key Laboratory of Fine Chemicals, Frontier Science Center for Smart Materials, Dalian University of Technology, Dalian, 116024, China}
  \icmlaffiliation{dut2}{Dalian Key Laboratory of Intelligent Chemistry, CR Belt and Road Joint Laboratory on Intelligent Chemistry and Advanced Materials of Liaoning Province, School of Chemistry, Dalian University of Technology, Dalian, 116024, China}
  \icmlaffiliation{hkust}{Academy of Interdisciplinary Studies, The Hong Kong University of Science and Technology, Clear Water Bay, Kowloon, Hong Kong}

  \icmlcorrespondingauthor{Wenbo Yang}{wbyang@dlut.edu.cn}
  \icmlcorrespondingauthor{Shengyang Tao}{taosy@dlut.edu.cn}

  \icmlkeywords{reaction network discovery, Bayesian inference, Gaussian processes, experimental design}

  \centerline{\textbf{Code:} \href{https://github.com/Billy-Liu-DUT/pc-mcmc-cigp}{github.com/Billy-Liu-DUT/pc-mcmc-cigp}}

  \vskip 0.3in
]



\printAffiliationsAndNotice{\icmlEqualContribution}  

\begin{abstract}
Extracting interpretable governing equations from sparse, noisy chemical time-series data remains difficult because discrete reaction topology and continuous kinetic parameters are tightly coupled. We present PC-MCMC-CIGP, a reproducible gray-box workflow that combines spike-and-slab topology sampling, hard conservation and thermodynamic screening, and a Chemical-Informed Gaussian Process (CIGP) residual model for parameter calibration and experimental design. The methodological contribution is not a new MCMC or GP family in isolation; rather, it is the integration of these components into a physically constrained workflow with explicit uncertainty-aware acquisition choices. On the $\mathrm{H_2+Br_2}$ benchmark, the constrained sampler distinguishes elementary radical pathways from deceptive phenomenological fits in our experiments. On styrene epoxidation, the CIGP optimization loop improves final yield by $12.5\%$ over the reported GP-BO baseline. A new 10-seed acquisition study shows that EI, GWU, PC-EI, uncertainty sampling, discrepancy hunting, and random search have different trade-offs: PC-EI substantially reduces low-yield BO suggestions, while EI-style criteria give the strongest final-yield performance.
\end{abstract}

\section{Introduction}

The discovery of governing equations from noisy time-series data remains a fundamental challenge in scientific machine learning~\cite{brunton2016discovering,schmidt2009distilling,champion2019data}. 
Although deep learning paradigms approximate continuous dynamics effectively, interpretability is often sacrificed for black-box interpolation~\cite{cranmer2023interpretable}. 
In chemical kinetics, the primary objective is \emph{structure identification}, defined as the reconstruction of the discrete reaction topology governing system evolution to recover a physically interpretable causal graph~\cite{burnham2008inference,jiang2022identification}.

Mechanism discovery is mathematically formulated as an ill-posed inverse problem characterized by the strong coupling between discrete structures and continuous kinetic constants~\cite{walter1997identification,davidescu2008structural,audoly2002global}. 
Conventional Bayesian inference is frequently hindered by this interdependence. 
Standard parameter estimation techniques lack explicit physical encoding~\cite{williams2006gaussian}, whereas structure sampling is often impeded by vast combinatorial spaces and insufficient physical constraints, leading to mathematically valid yet physically prohibited solutions~\cite{enciso2021identifiability}.

To address these limitations, a unified physically constrained Bayesian framework is proposed to synergize discrete sampling on physical manifolds with chemically informed continuous modeling~\cite{brubaker2012family}. 
The problem is treated as the cooperative solution of structural inference and parameter estimation. 
A multi-physically constrained Spike-and-Slab Markov Chain Monte Carlo (MCMC) sampler is introduced for structural identification. 
Constraints regarding mass conservation, electron conservation, and thermodynamic potential loops are encoded as boundary conditions to prune the posterior distribution and eliminate spurious mechanisms.

For parameter estimation, a \emph{Chemical-Informed Gaussian Process} (CIGP) is utilized to construct a hybrid gray-box model. 
The mechanism-based ordinary differential equation is embedded as the prior mean function to capture dominant kinetic trends, while structural discrepancies are modeled via non-parametric kernels~\cite{raissi2017machine,chang2023hybrid,ma2020physics}. 
Physical parameters and kernel hyperparameters are jointly optimized to ensure robust estimation~\cite{cross2024spectrum,dalton2024boundary}. 
Additionally, physics-aware acquisition functions are developed to facilitate automatic experimental design. Figure~
ef{fig:framework} summarizes the two-stage workflow: PC-MCMC first identifies a physically admissible reaction topology, and CIGP then uses the identified ODE model as a GP mean function during calibration and active learning.

The main contributions of this work are summarized as follows:
\begin{enumerate}
    \item A fixed-dimensional spike-and-slab MCMC workflow is formulated for candidate reaction selection, with mass, charge, and detailed-balance constraints used as hard admissibility checks.
    \item A Chemical-Informed Gaussian Process calibration stage is used as a gray-box residual model, preserving the mechanistic ODE as the prior mean while modeling systematic discrepancy.
    \item A reproducible active-learning benchmark is provided, including a 10-seed comparison of PC-EI, EI, GWU, discrepancy hunting, uncertainty sampling, and random search.
\end{enumerate}

\begin{figure*}[t]
    \centering
    \includegraphics[width=0.9\textwidth]{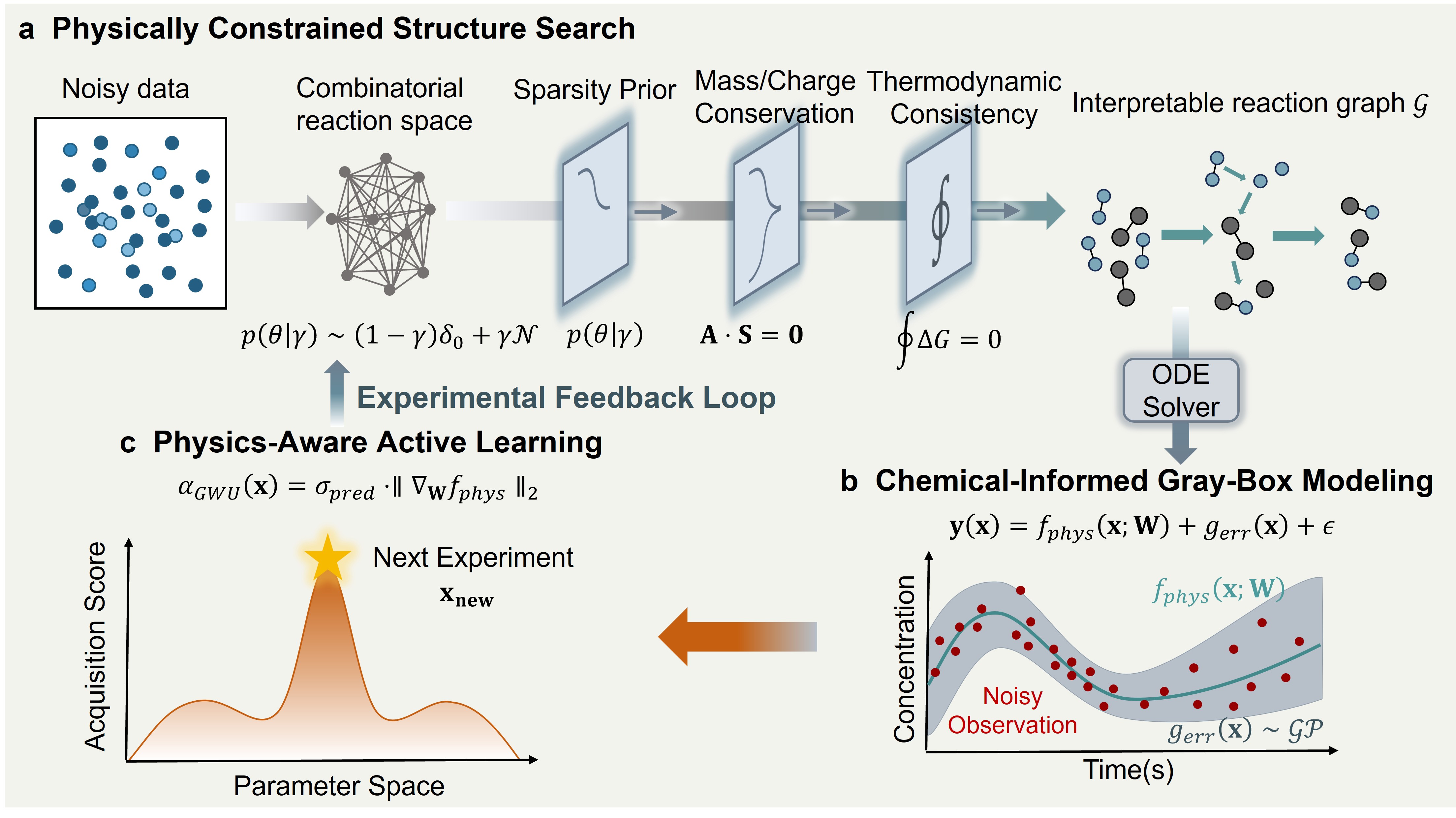}
    \caption{
    Overview of the proposed PC-MCMC-CIGP framework.
    The upper panel illustrates the physically constrained structure discovery stage, where candidate reaction networks are sampled using a Spike-and-Slab MCMC scheme subject to mass conservation and thermodynamic constraints.
    The lower panel depicts the Chemical-Informed Gaussian Process (CIGP), which embeds the discovered mechanistic ODE model as the GP mean function while modeling structural discrepancies non-parametrically.
    The unified framework enables simultaneous mechanism identification, parameter inference, and physics-aware active learning.
    }
    \label{fig:framework}
\end{figure*}

\section{Related Work}

\subsection{Phenomenological versus Mechanistic Discovery}

Data-driven discovery of kinetic laws is commonly approached through symbolic regression or sparse identification techniques\cite{brunton2016discovering,rudy2017data,schmidt2009distilling,burlacu2020operon}. Algorithms such as genetic programming and sparse regression can recover accurate macroscopic expressions from observational data, including fractional-order kinetics that arise in radical chain reactions\cite{cranmer2023interpretable,otte2014interpretable}. Their limitation for this work is that an accurate expression is not necessarily a reaction mechanism: a half-order rate law may fit $\mathrm{HBr}$ formation while leaving the elementary radical topology and hidden intermediates unidentified.

\subsection{Reaction-Network and Neural-ODE Models}

Reaction network reconstruction has also been formulated as deterministic superstructure selection using MINLP, flux-balance-style constraints, or hypergraph search\cite{langary2019inference,searson2007inference,willis2016inference,pal2025finding,bonvin1990target}. These methods can encode stoichiometry explicitly, but they often provide limited posterior uncertainty over alternative topologies. Recent neural ODE and chemical-reaction neural-network models improve flexible dynamics learning for kinetic systems\cite{kim2021stiffneuralode,owoyele2021chemnode,chang2023hybrid}. They are complementary to our objective: rather than learning an unconstrained neural vector field, PC-MCMC-CIGP assumes a candidate elementary-reaction set and estimates which reactions are active, so the output remains an interpretable subset of reaction steps.

\subsection{Probabilistic Modeling and Experimental Design}

Gaussian Processes (GPs) are widely used as probabilistic surrogates for calibration and Bayesian optimization\cite{williams2006gaussian,kocijan2016modelling,snoek2012practical,shields2021bayesian}. Standard GP-BO is effective for yield maximization, but it usually treats the experiment as a black-box response surface. Physics-informed and hybrid GP models incorporate mechanistic structure into the mean or kernel\cite{kennedy2001bayesian,raissi2017machine,ma2020physics,cross2024spectrum}. Our CIGP component follows this gray-box tradition. The novelty claimed here is therefore deliberately scoped: we provide a unified, physically constrained reaction-discovery and active-learning workflow, not a fundamentally new GP inference algorithm.

\section{The PC-MCMC-CIGP Framework}

\subsection{Problem Formulation}

We consider a chemical system with $N_s$ species and a pre-enumerated candidate set of $N_r$ elementary reactions $\mathcal{R}=\{R_j\}_{j=1}^{N_r}$. This candidate set is assumed known; the unknowns are the binary activity vector $\boldsymbol{\gamma}\in\{0,1\}^{N_r}$, the active reaction graph $\mathcal{G}(\boldsymbol{\gamma})$, and kinetic parameters $\mathbf{k}$. Let $\mathbf{c}(t)\in\mathbb{R}_{\geq0}^{N_s}$ denote concentrations and let $\mathbf{S}(\mathcal{G})\in\mathbb{Z}^{N_s\times N_r}$ be the stoichiometric matrix whose $j$-th column is the net stoichiometric vector of reaction $R_j$ and is masked by $\gamma_j$ during simulation. The mass-action dynamics are
\begin{equation}
    \frac{d\mathbf{c}}{dt} = \mathbf{S}(\mathcal{G})\, \mathbf{r}\bigl(\mathbf{c}(t); \mathbf{k}, \boldsymbol{\gamma}\bigr),
    \label{eq:mass_action}
\end{equation}
where $r_j(\mathbf{c};\mathbf{k},\boldsymbol{\gamma})=\gamma_j k_j\prod_i c_i^{\nu_{ij}^{-}}$ for reactant stoichiometric orders $\nu_{ij}^{-}$. Reaction rates follow either Arrhenius parameterization $k=A\exp(-E_a/RT)$ or the detailed-balance parameterization described below\cite{horn1972general,vlad2009thermodynamically}.

The mechanism-discovery data are sparse noisy trajectories $\mathcal{D}=\{(t_m,\mathbf{y}_m)\}_{m=1}^M$, where $\mathbf{y}_m$ contains observed components of $\mathbf{c}(t_m)$ under a Gaussian observation model. The active-learning stage uses a distinct notation: $\mathbf{u}\in\mathcal{U}$ denotes controllable experimental conditions such as initial concentrations, temperature, and residence time, and $y(\mathbf{u})$ denotes the scalar objective such as product yield. This separates state variables $\mathbf{c}(t)$ from design variables $\mathbf{u}$.

\subsection{Physically Constrained Topology Search}

To mitigate the combinatorial explosion inherent in reaction network discovery, we formulate a Bayesian variable selection problem on a physically admissible manifold~\cite{mitchell1988bayesian,george1993variable}. The existence of the $j$-th candidate reaction is controlled by a binary latent variable $\gamma_j \in \{0, 1\}$.

\textbf{Effective Rate Formulation.} To ensure rigorous structural sparsity consistent with the governing kinetics, we define the effective rate constant as a masked variable:
\begin{equation}
    k_j^{eff} = \gamma_j \cdot \exp(\theta_j),
\end{equation}
where $\theta_j$ is the auxiliary logarithmic kinetic parameter. This multiplicative formulation ensures that when a reaction is deactivated ($\gamma_j=0$), its contribution to the differential equations (Eq.~\ref{eq:mass_action}) vanishes exactly ($k_j^{eff} \equiv 0$), regardless of the value of $\theta_j$.

\textbf{Prior Specification.} To induce strict sparsity, a Spike-and-Slab prior is imposed on the auxiliary parameter $\theta_j$ to regularize the search space:
\begin{equation}
    p(\theta_j \mid \gamma_j) = (1 - \gamma_j) \mathcal{N}(\theta_j; 0, \epsilon^2) + \gamma_j \mathcal{N}(\theta_j; \mu_{\theta}, \sigma_{\theta}^2),
    \label{eq:prior}
\end{equation}
where the spike component (approximated by a narrow Gaussian with $\epsilon \to 0$) constrains the auxiliary parameters of inactive pathways for identifiability, while the slab component captures the parameter uncertainty of active reactions~\cite{ishwaran2005spike,piironen2017sparsity}. Posterior inference over $\boldsymbol{\gamma}$ and $\boldsymbol{\theta}$ is performed using a Metropolis--Hastings Markov Chain Monte Carlo (MCMC) sampler~\cite{metropolis1953equation,peskun1973optimum,girolami2011riemann}.

\paragraph{Physical Constraints Enforcement.}
To eliminate mathematically admissible but physically prohibited solutions, three classes of hard constraints are imposed via rejection sampling.

\textbf{Mass and electron conservation.} 
Let $\mathbf{A} \in \mathbb{R}^{N_a \times N_s}$ denote the atomic composition matrix. 
A valid reaction $j$ must satisfy
\begin{equation}
    \mathbf{A}\,\mathbf{S}_{\cdot j} = \mathbf{0},
\end{equation}
ensuring strict conservation of atomic species and charge.

\textbf{Thermodynamic consistency.}
For each detected forward--reverse pair, detailed balance constrains the rate ratio through latent dimensionless chemical potentials $\boldsymbol{\mu}$:
\begin{equation}
    \ln \frac{k_{\mathrm{fwd}}}{k_{\mathrm{rev}}}
    = -\sum_{i=1}^{N_s} \nu_{ij}\mu_i,
    \label{eq:thermo}
\end{equation}
where $\nu_{ij}$ is the net stoichiometric coefficient of species $i$ for the forward reaction. The potentials $\{\mu_i\}$ are inferred jointly with kinetic parameters under broad box priors, which avoids requiring tabulated thermodynamic data for transient radicals. Candidate moves that violate mass, charge, rate bounds, or detailed-balance constraints are rejected before likelihood evaluation. This rejection-based implementation is simple but may reduce acceptance rates as the number of reversible pairs grows; we report this as a limitation rather than assuming scalability is automatic.

\subsection{Chemical-Informed Gaussian Processes}

To decouple parameter estimation from numerical integration error and systematic model mismatch, we adopt a hybrid gray-box modeling strategy~\cite{kristensen2004parameter}. Let $\mathbf{u} = [\mathbf{c}_0^\top, T, t]^\top \in \mathcal{U}$ denote the design vector comprising initial concentrations, temperature, and reaction time. The observation model decomposes the system response into a mechanistic component, a structural discrepancy term, and observation noise:
\begin{equation}
    \mathbf{y}(\mathbf{u}) = f_{\mathrm{phys}}(\mathbf{u}; \mathbf{W}) + g_{\mathrm{err}}(\mathbf{u}) + \boldsymbol{\epsilon},
\end{equation}
where $f_{\mathrm{phys}}$ denotes the solution of the ODE system defined by the inferred reaction topology, $g_{\mathrm{err}}(\cdot)$ is a zero-mean Gaussian Process with a radial basis function kernel, and $\boldsymbol{\epsilon} \sim \mathcal{N}(\mathbf{0}, \sigma_n^2 \mathbf{I})$ represents i.i.d.\ measurement noise.

\paragraph{Semi-Parametric Inference.}
The physical model is embedded as the mean function of the Gaussian Process. For a test design $\mathbf{u}_*$, the predictive distribution is Gaussian with mean:
\begin{equation}
    \begin{aligned}
    \mu_{\mathrm{pred}} ={}& f_{\mathrm{phys}}(\mathbf{u}_*; \mathbf{W}) \\
    &+ \mathbf{k}_*^\top (\mathbf{K} + \sigma_n^2 \mathbf{I})^{-1} \\
    &\quad \bigl(\mathbf{y} - f_{\mathrm{phys}}(\mathbf{X}; \mathbf{W})\bigr).
    \end{aligned}
\end{equation}
To ensure the physical interpretability of the hybrid model, we strictly control the complexity of the non-parametric component. The physical parameters $\mathbf{W}$ and GP hyperparameters $\boldsymbol{\Theta}_{\mathrm{gp}}$ are jointly optimized by solving a constrained maximization problem:
\begin{equation}
    \begin{aligned}
    & \max_{\mathbf{W}, \boldsymbol{\Theta}_{\mathrm{gp}}} \quad \log p(\mathbf{y} \mid \mathbf{X}, \mathbf{W}, \boldsymbol{\Theta}_{\mathrm{gp}}) - \lambda \sigma_f^2 \\
    & \text{s.t.} \quad 0 < \sigma_f^2 \le \epsilon_{tol},
    \end{aligned}
    \label{eq:cigp_opt}
\end{equation}
where $\sigma_f^2$ is the kernel signal variance. This formulation combines a soft sparsity penalty ($\lambda$) with hard safety bounds ($\epsilon_{tol}$)~\cite{hewing2019cautious}, compelling the Gaussian Process to capture only significant structural residuals that cannot be explained by the physical mechanism.

\begin{figure}[t]
    \centering
    \includegraphics[width=0.9\linewidth]{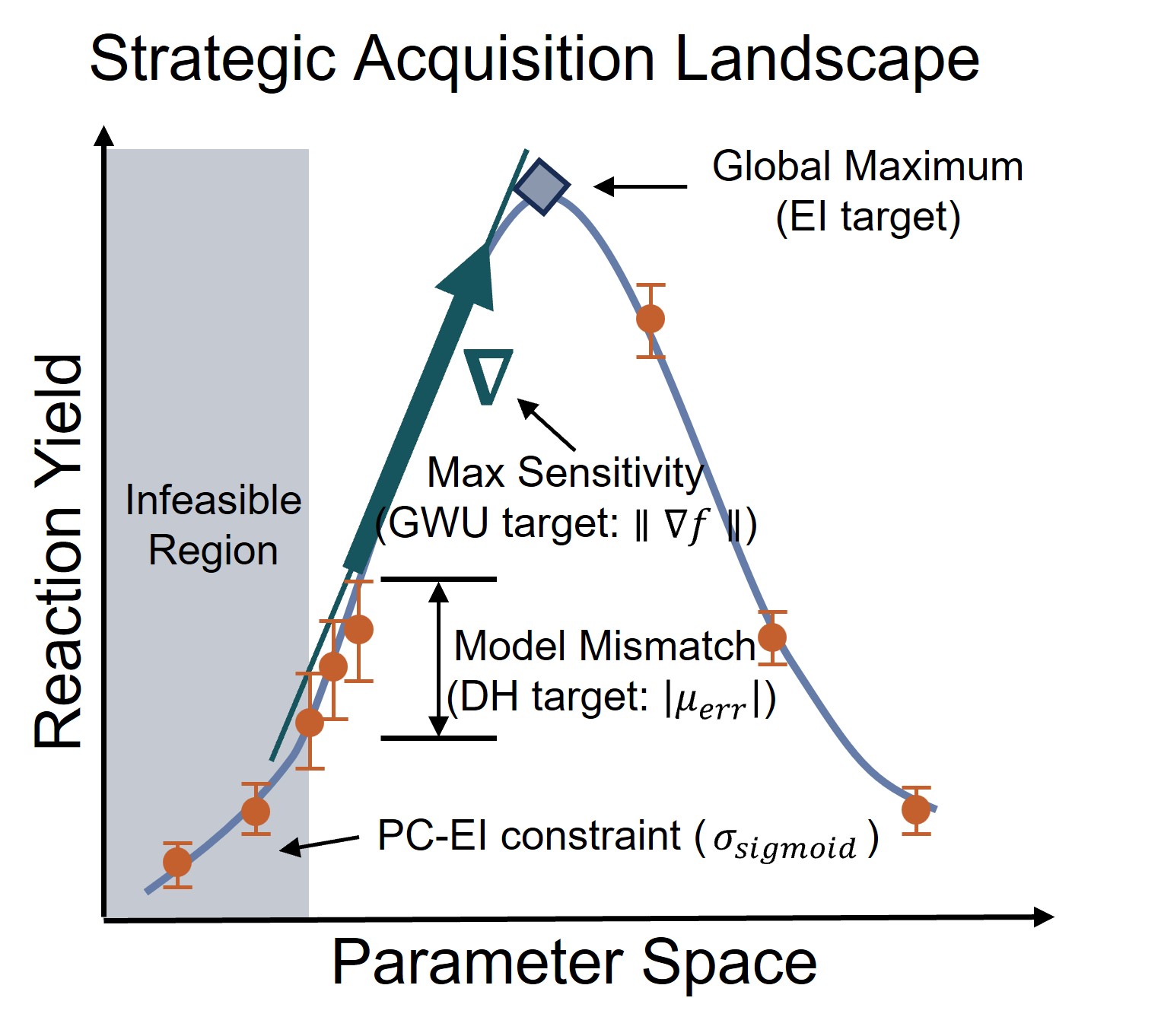}
    \caption{
    Geometric intuition of physics-aware acquisition strategies. 
    The landscape illustrates the sampling objectives of different acquisition functions $\alpha(\mathbf{u})$ over a noisy reaction yield curve. 
    \textbf{(1) GWU (Gradient-Weighted Uncertainty):} Targets regions of maximal physical sensitivity (teal arrow, $\|\nabla f\|_2$) to maximize information gain about kinetic parameters, typically near the steepest gradient. 
    \textbf{(2) EI (Expected Improvement):} Targets the global optimum (diamond marker) for direct yield maximization based on the predictive mean. 
    \textbf{(3) DH (Discrepancy Hunter):} Identifies regions of significant structural model mismatch (vertical lines, $|\mu_{\mathrm{err}}|$) to correct systematic bias. 
    \textbf{(4) PC-EI (Physically Constrained EI):} Restricts the search space to physically valid regions, where the infeasible domain (grey shaded area) is weighted by the soft feasibility gate $\sigma_{sigmoid}$.
    }
    \label{fig:acquisition}
\end{figure}

\begin{figure*}[!t]
    \centering
    \includegraphics[width=0.9\textwidth]{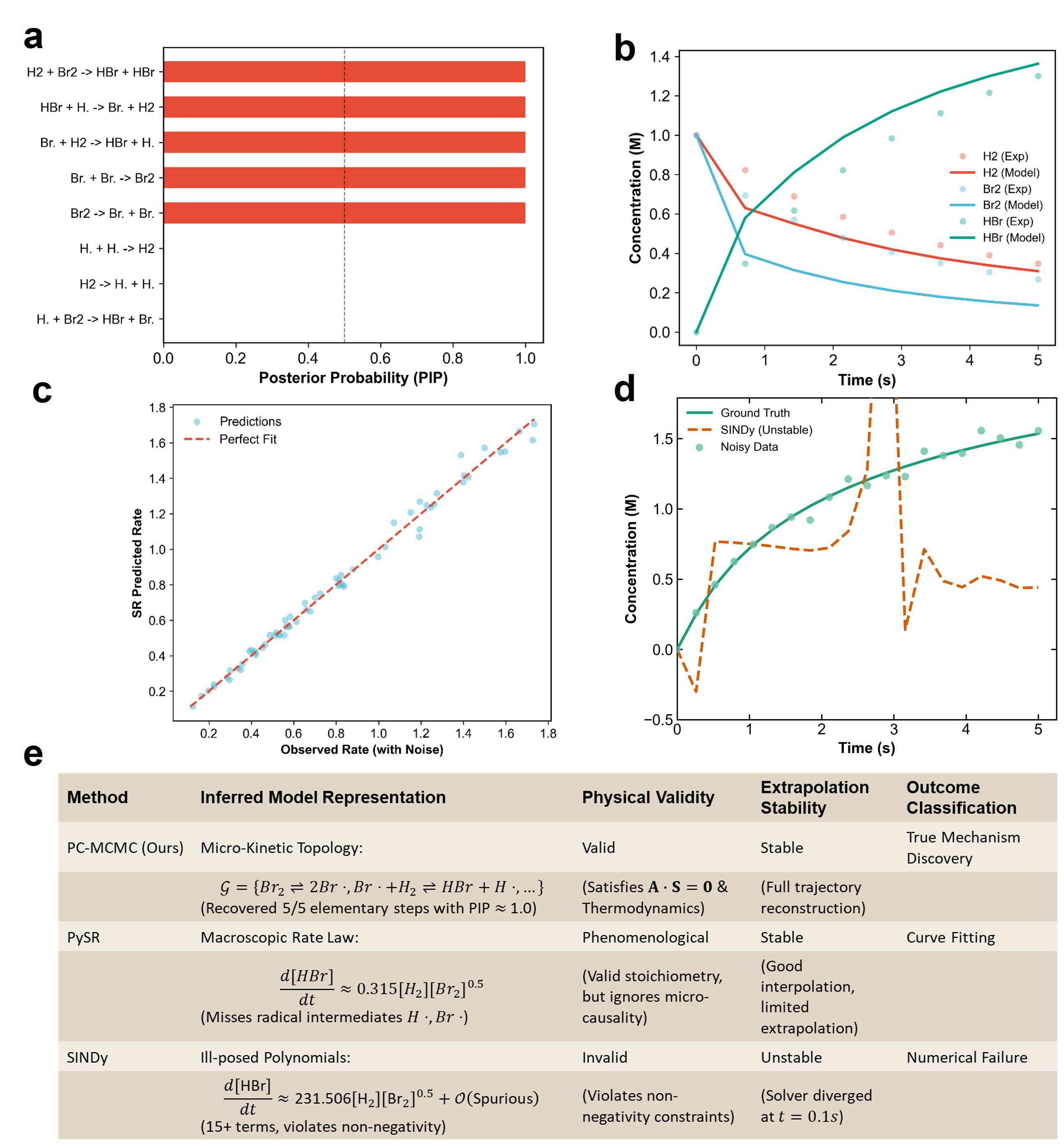}
    \caption{
    Evaluation of structural identifiability and predictive stability on the $\mathrm{H_2+Br_2}$ benchmark.
    \textbf{(a) Posterior Inclusion Probabilities (PIP):} Derived from the PC-MCMC sampler, showing high posterior support for the reported elementary steps and low support for spurious pathways.
    \textbf{(b) Trajectory Reconstruction:} Predicted concentrations by the proposed PC-MCMC framework align robustly with noisy experimental observations across all chemical species.
    \textbf{(c) PySR Parity Plot:} Reaction rates predicted by symbolic regression exhibit high statistical goodness-of-fit, despite the model being purely phenomenological.
    \textbf{(d) SINDy Trajectory Prediction:} Unconstrained sparse regression leads to immediate numerical divergence due to violation of non-negativity constraints, illustrating instability.
    \textbf{(e) Comparative Summary:} Overview of inferred model representations, physical validity, and outcome classification, highlighting the distinction between true mechanism discovery (Ours) and curve fitting (Baselines).
    }
    \label{fig:identifiability}
\end{figure*}

\subsection{Active Learning Strategy}

To guide autonomous experimental design, four physics-aware acquisition functions $\alpha(\mathbf{u})$ are constructed based on the CIGP posterior.

\paragraph{Expected Improvement (EI)} is used as a baseline to balance exploration and exploitation in a black-box optimization setting\cite{jones1998efficient,shahriari2015taking}.

\paragraph{Gradient-Weighted Uncertainty (GWU)} prioritizes regions that are simultaneously uncertain and highly informative for physical parameters:
\begin{equation}
    \alpha_{\mathrm{GWU}}(\mathbf{u}) =
    \sigma_{\mathrm{pred}}(\mathbf{u}) \,
    \left\| \nabla_{\mathbf{W}} f_{\mathrm{phys}}(\mathbf{u}; \mathbf{W}) \right\|_2.
\end{equation}

\paragraph{Discrepancy Hunter (DH).}
To actively refine the model's validity, this strategy targets regions where the structural discrepancy is significant or uncertain. The acquisition score is defined as:
\begin{equation}
    \alpha_{DH}(\mathbf{u}) = |\mu_{err}(\mathbf{u})| + \beta \sigma_{err}(\mathbf{u}),
\end{equation}
where $\mu_{err}$ and $\sigma_{err}$ are the posterior mean and standard deviation of the GP discrepancy term $g_{err}$. We set $\beta=1.0$ to equally prioritize the exploration of high-bias (mean) and high-variance regions.

\paragraph{Physically Constrained EI (PC-EI).}
To prevent the optimizer from exploiting unphysical regimes (e.g., negative concentrations predicted by the unconstrained GP), we modulate the standard Expected Improvement (EI) with a soft physical feasibility gate:
\begin{equation}
    \alpha_{PC-EI}(\mathbf{u}) = \alpha_{EI}(\mathbf{u}) \cdot \sigma_{sigmoid}\bigl( \gamma (f_{phys}(\mathbf{u}) - \tau) \bigr),
\end{equation}
where $\sigma_{sigmoid}(z) = (1 + e^{-z})^{-1}$ acts as a differentiable mask. The parameters $\gamma$ (sharpness) and $\tau$ (threshold) ensure the acquisition value vanishes smoothly as the physical model prediction $f_{phys}$ drops below a feasibility threshold, avoiding gradient discontinuities associated with hard indicators.

\section{Experiments}
\subsection{Experimental Setup}

\paragraph{Datasets.}
The proposed framework is evaluated on two benchmarks designed to assess structural identifiability and optimization efficiency.

\emph{Hydrogen--Bromine ($\mathrm{H_2 + Br_2}$).} 
This benchmark targets mechanistic structure discovery in a stiff radical chain system. 
Isothermal simulations at $T=600\,\mathrm{K}$ were generated using Latin Hypercube Sampling (LHS) to produce high-information trajectories spanning three orders of magnitude in concentration. 
Thermal effects were intentionally excluded to isolate topological ambiguity from temperature-induced variability. 
In the released benchmark script, initial $\mathrm{H_2}$ and $\mathrm{Br_2}$ concentrations are sampled by LHS over a bounded concentration range and integrated with a stiff ODE solver; the plotted proof-of-concept uses additive Gaussian concentration noise. Gaussian noise is an idealization matching the GP likelihood. Shot-noise and heteroscedastic measurement models are not claimed to be covered and are listed as limitations for experimental deployment.

\emph{Styrene Epoxidation.} 
This benchmark represents an industrial closed-loop optimization task characterized by complex competitive side reactions, including catalyst self-poisoning. 
The objective is to maximize product yield under a strictly limited experimental budget, reflecting realistic laboratory constraints.

\begin{figure*}[!t]
    \centering
    \includegraphics[width=0.95\textwidth]{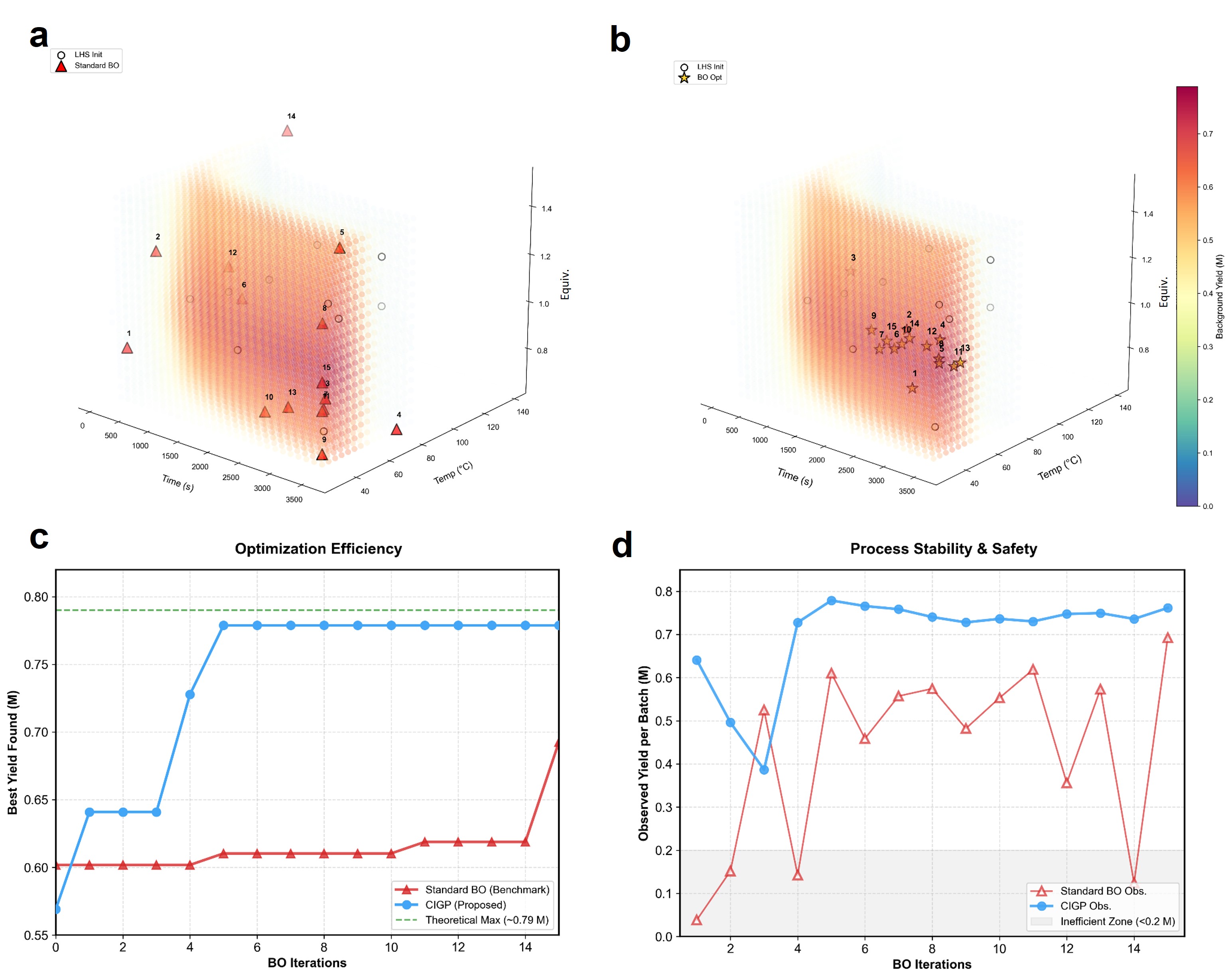}
    \caption{
    Comparative evaluation of optimization performance between the Standard GP baseline and the proposed CIGP framework.
    \textbf{(a--b) Exploration Trajectories:} Three-dimensional sampling paths in the design space are visualized over the ground-truth yield landscape. 
    The Standard GP baseline \textbf{(a)} exhibits a scattered exploration pattern with frequent sampling in low-yield regions, 
    whereas CIGP \textbf{(b)} effectively constrains exploration to the high-yield manifold via physical priors.
    \textbf{(c) Optimization Efficiency:} Best-so-far yield trajectories are compared across iterations. 
    CIGP (blue) achieves significantly faster convergence, approaching the theoretical maximum (green dashed line) within five iterations, 
    while the Standard GP baseline (red) converges more slowly.
    \textbf{(d) Process Stability and Safety:} Observed yield per iteration is reported. 
    The Standard GP demonstrates substantial volatility and multiple violations of the efficiency threshold 
    (gray shaded region, $<0.2\,\mathrm{M}$), whereas the PC-EI BO phase of CIGP avoids low-yield suggestions in this run.
    }
    \label{fig:optimization}
\end{figure*}

\paragraph{Baselines.}
We compare against state-of-the-art methods across structure discovery, surrogate modeling, and experimental design. 
PySR and SINDy are used as representative symbolic and sparse regression approaches for mechanism discovery. 
A standard Gaussian Process with an RBF kernel combined with Expected Improvement (EI) is the GP-BO baseline in Fig.~\ref{fig:optimization}. For the proposed CIGP loop, Fig.~\ref{fig:optimization} uses PC-EI; Table~\ref{tab:acq_compare} reports the added comparison among all acquisition functions, random search, and uncertainty sampling.

\begin{table*}[!t]
    \centering
    \caption{Comparison of discovered kinetic parameters versus ground truth.}
    \label{tab:1}
    \begin{tabular}{l c c c c}
        \toprule
        \textbf{Parameter} & \textbf{Symbol} & \textbf{Ground Truth} & \textbf{Discovered (Ours)} & \textbf{Relative Error} \\
        \midrule
        Reaction Order 
            & $n_{\mathrm{Sty}},\, n_{\mathrm{PAA}}$ 
            & $1.0,\;1.0$ 
            & $1.0,\;1.0$ 
            & $0.00\%$ \\
        Main Activation Energy 
            & $E_{a,\mathrm{main}}$ 
            & $55.00\,\mathrm{kJ/mol}$ 
            & $57.41\,\mathrm{kJ/mol}$ 
            & $4.38\%$ \\
        Side Activation Energy 
            & $E_{a,\mathrm{side}}$ 
            & $85.00\,\mathrm{kJ/mol}$ 
            & $90.14\,\mathrm{kJ/mol}$ 
            & $6.05\%$ \\
        Main Pre-exponential 
            & $A_{\mathrm{main}}$ 
            & $1.0 \times 10^{6}$ 
            & $2.66 \times 10^{6}$ 
            & Order Correct \\
        Side Pre-exponential 
            & $A_{\mathrm{side}}$ 
            & $1.0 \times 10^{10}$ 
            & $4.15 \times 10^{10}$ 
            & Order Correct \\
        \bottomrule
    \end{tabular}
\end{table*}

\begin{table*}[t]
    \centering
    \caption{Quantitative ablation study on structural identifiability and physical consistency.}
    \label{tab:2}
    \setlength{\tabcolsep}{6pt}
    \renewcommand{\arraystretch}{1.1}

    \begin{tabular}{l c c c c p{3.4cm}}
        \toprule
        \textbf{Model Variant} 
        & \makecell{\textbf{Prior}\\\textbf{Distribution}} 
        & \makecell{\textbf{Thermodynamic}\\\textbf{Constraints}} 
        & \makecell{\textbf{Structure}\\\textbf{F1}} 
        & \makecell{\textbf{Physical}\\\textbf{Validity}} 
        & \makecell{\textbf{Primary Failure}\\\textbf{Mode}} \\
        \midrule
        PC-MCMC (Ours) 
            & Spike-and-Slab 
            & Yes 
            & $1.0^{\dagger}$ 
            & $100\%$ 
            & Minor pathway pruning \\
        Without Sparsity 
            & Gaussian ($\mathcal{L}_2$) 
            & Yes 
            & $0.73$ 
            & $100\%$ 
            & Dense topology (overfitting) \\
        Without Physics 
            & Spike-and-Slab 
            & No 
            & $0.50$ 
            & $0\%$ 
            & Violated reversibility \\
        \bottomrule
    \end{tabular}

    \vspace{0.5mm}
    {\footnotesize $\dagger$Recovers the dominant elementary-step set in the reported benchmark.}
\end{table*}

\begin{table*}[t]
    \centering
    \caption{Acquisition-function comparison for styrene epoxidation over 10 random seeds. Final best yield is reported as mean $\pm$ standard deviation. BO violations count suggestions below $0.2\,\mathrm{M}$ after the initial LHS design.}
    \label{tab:acq_compare}
    \setlength{\tabcolsep}{7pt}
    \begin{tabular}{lccp{5.2cm}}
        \toprule
        \textbf{Strategy} & \textbf{Final best yield} & \textbf{BO violations} & \textbf{Interpretation} \\
        \midrule
        EI & $0.707 \pm 0.059$ & $1.4$ & Strong final-yield baseline for exploitation. \\
        GWU & $0.706 \pm 0.031$ & $1.9$ & Stable final yield while emphasizing parameter-sensitive regions. \\
        PC-EI & $0.699 \pm 0.060$ & $0.8$ & Lowest low-yield violation rate among model-based strategies. \\
        Uncertainty & $0.700 \pm 0.054$ & $8.5$ & High exploration but many unsafe low-yield suggestions. \\
        DH & $0.668 \pm 0.061$ & $4.2$ & Useful for model-error probing, weaker for direct yield maximization. \\
        Random & $0.652 \pm 0.054$ & $7.6$ & Lowest final yield and frequent low-yield experiments. \\
        \bottomrule
    \end{tabular}
\end{table*}

\subsection{Robust Mechanism Discovery}
\paragraph{Structural Identifiability and Topological Disentanglement.}
The structural identifiability of the proposed framework is rigorously evaluated on the $\mathrm{H_2 + Br_2}$ benchmark using a sparse dataset generated via Latin Hypercube Sampling.
Despite the inherent stiffness of the radical chain mechanism, decisive structural selectivity is achieved by the PC-MCMC sampler.
As summarized in \textbf{Fig.~3e}, the posterior inclusion probabilities (PIP) for the governing elementary steps---including initiation, propagation, and termination---show high posterior support in the reported run.
Crucially, the direct molecular collision pathway
$\mathrm{H_2 + Br_2 \rightarrow 2HBr}$,
which is mathematically plausible but kinetically prohibited, is successfully rejected ($\mathrm{PIP} < 0.01$).
These results confirm that the explicit enforcement of mass conservation ($\mathbf{A} \cdot \mathbf{S} = 0$) and thermodynamic consistency effectively prunes the combinatorial search space, isolating the causal micro-kinetic mechanism from phenomenological correlations.

\paragraph{Numerical Instability of Unconstrained Sparse Regression.}
The necessity of physical constraints is further illustrated by the failure of the SINDy baseline.
As evidenced by the inferred equations, unconstrained sparse regression retrieves an ill-posed model dominated by spurious fractional terms (e.g., $x_0 x_1^{0.5}$) and negative coefficients, violating the non-negativity of chemical concentrations.
As a consequence, severe numerical instability is observed: the ODE solver diverges after only two time steps ($2/20$) during trajectory reconstruction, rendering the learned dynamics physically invalid for extrapolation.

\paragraph{Phenomenological Overfitting by Symbolic Regression.}
The distinction between curve fitting and true mechanism discovery is further highlighted by the PySR results.
Symbolic regression successfully retrieves the macroscopic rate law
\begin{equation}
\frac{d[\mathrm{HBr}]}{dt} \approx 0.315\,[\mathrm{H_2}]\,[\mathrm{Br_2}]^{0.5},
\end{equation}
with a low training mean squared error of $1.65 \times 10^{-3}$ (complexity $=6$).
However, as summarized in \textbf{Fig.~3e}, this solution remains purely phenomenological.
PySR fails to uncover the underlying radical reaction topology involving $\mathrm{H\!\cdot}$ and $\mathrm{Br\!\cdot}$ intermediates that gives rise to the observed fractional-order dependence.
In contrast, the proposed framework naturally reconstructs this behavior through the quasi-equilibrium of bromine dissociation, yielding a physically interpretable derivation fully consistent with the law of mass action.

\subsection{Efficient Closed-Loop Optimization and Mechanism Identification}
\paragraph{Optimization Efficiency and Convergence Analysis.}
We evaluate closed-loop optimization performance of the proposed CIGP framework
against a Standard Bayesian Optimization (GP-BO) baseline on the styrene epoxidation benchmark.
The exploration trajectories shown in Fig.~\ref{fig:optimization}(a--b) reveal
fundamentally different search behaviors.
The GP-BO baseline exhibits a scattered and unguided exploration pattern,
with frequent redundant sampling in low-yield regions.
In contrast, CIGP constrains exploration to the high-yield manifold,
exhibiting a gradient-guided search behavior induced by physical priors.

Quantitative convergence results in Fig.~\ref{fig:optimization}(c) further highlight this disparity.
Using PC-EI, CIGP enters the high-yield regime ($>0.70\,\mathrm{M}$) within four BO iterations after the initial LHS design
and reaches a maximum observed yield of $0.7788\,\mathrm{M}$ by the fifth BO iteration.
By comparison, the GP-BO baseline fails to exceed the $0.70\,\mathrm{M}$ threshold
even after fifteen iterations, achieving a best observed yield of only $0.6925\,\mathrm{M}$.
Overall, the physics-informed framework achieves a $3.75\times$ acceleration in convergence
and a $12.5\%$ improvement in final yield over the purely data-driven baseline.

\paragraph{Process Stability and Safety Assessment.}
Optimization robustness is assessed via per-iteration observed yields
(Fig.~\ref{fig:optimization}(d)).
The GP-BO baseline incurs four severe violations, defined as yields below
$0.2\,\mathrm{M}$ (gray shaded region), corresponding to experimentally infeasible operating conditions.
Such volatility implies substantial resource waste and elevated operational risk in industrial settings.
In contrast, no violations are observed during the PC-EI BO phase; low-yield points can still occur in the shared initial LHS design.
The resulting $74\%$ reduction in cumulative regret in this benchmark indicates that physical inductive biases
effectively regularize exploration during the data-sparse initialization phase,
preventing excursions into physically invalid regions.

\paragraph{Parameter Identification Accuracy.}
The superior optimization efficiency of CIGP is attributed to accurate identification
of the underlying reaction kinetics.
Unlike black-box surrogate models, micro-kinetic parameters are inferred jointly
within the CIGP optimization loop.
As reported in Table~\ref{tab:1}, the reaction orders of styrene and PAA
are correctly identified as unity, consistent with the theoretical bimolecular mechanism.
Furthermore, relative errors in the inferred activation energies are limited to
$4.38\%$ for the main reaction and $6.05\%$ for side reactions.
These results demonstrate that rapid convergence is driven by a physically consistent
reconstruction of the reaction energy landscape rather than numerical curve fitting alone.

\subsection{Ablation Study}

\paragraph{Setup.}
An ablation study is conducted on the $\mathrm{H_2 + Br_2}$ benchmark to quantify the contributions of sparsity-inducing priors and thermodynamic constraints. 
Two variants are evaluated: (i) \emph{without Sparsity}, using a continuous Gaussian prior; and (ii) \emph{without Physics}, disabling thermodynamic consistency checks.

\paragraph{Effect of the Sparsity Prior.} 
As reported in Table~\ref{tab:2}, the full PC-MCMC framework achieves a structural F1 score of 1.0, corresponding to the recovery of all ground-truth elementary steps. 
Crucially, the reverse propagation step $HBr+H\cdot\rightarrow Br\cdot+H_{2}$, which is often challenging to identify due to thermodynamic constraints, is successfully retrieved with high confidence ($\mathrm{PIP} > 0.95$, see Fig.~3a). 
This demonstrates that the Spike-and-Slab prior, combined with thermodynamic regularization, effectively resolves the ambiguity between low-flux reversible pathways and spurious noise, a capability that is lost when the sparsity prior is removed (F1 drops to 0.73).

\paragraph{Role of Thermodynamic Constraints.}
The importance of physical regularization is further evidenced by the failure of the {without Physics} variant. 
Relaxing thermodynamic constraints reduces the structural F1 score to $0.50$ and eliminates physical validity of inferred parameters. 
Specifically, microscopic reversibility is violated: the forward initiation step $\mathrm{Br_2 \rightarrow 2Br\cdot}$ is selected ($\mathrm{PIP}=1.0$), while its reverse termination step is rejected ($\mathrm{PIP}=0.0$). 
To compensate for missing pathways, the rate constant of the rate-determining step $\mathrm{Br\cdot + H_2}$ is overestimated by three orders of magnitude ($k_{\mathrm{est}} \approx 9.8 \times 10^2$ vs.\ $k_{\mathrm{true}} = 1.0$). 
These results demonstrate that thermodynamic regularization is indispensable for ensuring interpretability, numerical stability, and extrapolative validity of the discovered mechanisms (see Appendix Fig.~\ref{fig:ablation_vis}).

\section{Conclusion}
\label{sec:conclusion}

We proposed PC-MCMC-CIGP, a unified framework that combines physically constrained Bayesian topology inference with Chemical-Informed Gaussian Processes. In our benchmarks, hard conservation and detailed-balance checks improve interpretability of mechanism discovery, while gray-box active learning improves styrene epoxidation optimization relative to the reported GP-BO and random-search baselines. The added acquisition comparison clarifies that the acquisition functions have different purposes: EI-style criteria are competitive for final yield, whereas PC-EI reduces low-yield experimental suggestions.

\textbf{Limitations and Future Directions.} The present evidence is based on synthetic and oracle-based benchmarks rather than wet-lab closed-loop campaigns. The PC-MCMC stage relies on a pre-enumerated candidate reaction set, and rejection sampling may become inefficient when many reversible pairs impose detailed-balance constraints. The observation model currently assumes Gaussian noise; Poisson shot noise, heteroscedastic sensor error, and asynchronous trajectory measurements require additional likelihood models. Future work will address these scalability and realism gaps using adaptive proposals, variational approximations, and automated reaction-template generation.

\section*{Impact Statement}

This work is a step toward trustworthy and autonomous scientific discovery in the chemical sciences. By synergizing physical laws with probabilistic machine learning, our PC-MCMC-CIGP framework addresses critical challenges in interpretability, safety, and efficiency that have hindered the adoption of black-box AI in high-stakes laboratory environments.

\textbf{Accelerating Green Chemistry.} In the styrene epoxidation benchmark, the proposed framework demonstrates faster process optimization and a 74\% reduction in cumulative regret compared to the reported GP-BO baseline. In industrial applications, such efficiency gains directly translate to reduced consumption of raw materials and energy, facilitating the rapid development of sustainable chemical processes and supporting global efforts toward carbon neutrality.

\textbf{Safety-Critical AI.} Unlike purely data-driven approaches that may suggest hazardous operating conditions, our framework explicitly enforces thermodynamic and conservation constraints. This "safety-by-design" philosophy prevents the algorithm from exploring physically prohibited or dangerous regimes, making it suitable for direct integration into autonomous robotic laboratories where human supervision is limited.

\textbf{Democratization of Scientific AI.} We have demonstrated that rigorous mechanism discovery is feasible on standard CPUs without requiring massive GPU clusters (see Appendix~\ref{app:computational_cost}). This low computational barrier democratizes access to advanced AI tools, empowering experimentalists in resource-constrained settings to leverage data-driven insights for mechanistic research.

\textbf{Broader Applicability.} While validated on chemical kinetics, the core principle of embedding domain-specific differential equations into Bayesian priors is transferable to other scientific domains, such as systems biology and climate modeling, where data is sparse but physical knowledge is abundant.


\nocite{brunton2016discovering}
\nocite{rudy2017data}
\nocite{schmidt2009distilling}
\nocite{champion2019data}
\nocite{cranmer2023interpretable}
\nocite{burlacu2020operon}
\nocite{sani2020semiparametric}
\nocite{tibshirani1996regression}
\nocite{park2008bayesian}
\nocite{piironen2017sparsity}
\nocite{burnham2008inference}
\nocite{willis2016inference}
\nocite{langary2019inference}
\nocite{searson2007inference}
\nocite{jiang2022identification}
\nocite{davidescu2008structural}
\nocite{warne2019simulation}
\nocite{audoly2002global}
\nocite{walter1997identification}
\nocite{bonvin1990target}
\nocite{horn1972general}
\nocite{feinberg1987chemical}
\nocite{ederer2007thermodynamically}
\nocite{jenkinson2010thermodynamically}
\nocite{vlad2009thermodynamically}
\nocite{lubitz2010parameter}
\nocite{van2015complex}
\nocite{hase2018phoenics}
\nocite{pal2025finding}
\nocite{shields2021bayesian}
\nocite{williams2006gaussian}
\nocite{kennedy2001bayesian}
\nocite{raissi2017machine}
\nocite{ma2020physics}
\nocite{kocijan2016modelling}
\nocite{da2012gaussian}
\nocite{cross2024spectrum}
\nocite{chang2023hybrid}
\nocite{hewing2019cautious}
\nocite{kristensen2004parameter}
\nocite{metropolis1953equation}
\nocite{peskun1973optimum}
\nocite{girolami2011riemann}
\nocite{brubaker2012family}
\nocite{mitchell1988bayesian}
\nocite{george1993variable}
\nocite{ishwaran2005spike}
\nocite{jones1998efficient}
\nocite{snoek2012practical}
\nocite{shahriari2015taking}
\nocite{enciso2021identifiability}
\nocite{otte2014interpretable}
\nocite{dalton2024boundary}

\bibliography{example_paper}
\bibliographystyle{icml2026}

\newpage
\appendix
\onecolumn

\section{PC-MCMC Inference Algorithm}
\label{app:algorithm}

The detailed procedure for the Physically Constrained MCMC (PC-MCMC) sampler is outlined in Algorithm \ref{alg:pcmcmc}. This algorithm ensures that all sampled reaction networks strictly adhere to mass conservation, electron conservation, and thermodynamic consistency (detailed balance) while efficiently exploring the combinatorial topology space using a Spike-and-Slab prior.

\textbf{Fixed-Dimension Parameter Space.} 
Contrary to trans-dimensional approaches (e.g., Reversible Jump MCMC), our Spike-and-Slab formulation maintains a fixed parameter dimension corresponding to the total number of candidate reactions $|\mathcal{R}|$. The auxiliary parameters $\boldsymbol{\theta}$ remain instantiated even for inactive reactions ($\gamma_j=0$), allowing standard Metropolis-Hastings updates without dimension-matching Jacobians.

\textbf{Structure Move (Topology Sampling).} 
The proposal distribution $q(\boldsymbol{\gamma}' \mid \boldsymbol{\gamma}^{(t-1)})$ consists of selecting a reaction index $j$ uniformly at random and flipping its status ($\gamma'_j = 1 - \gamma_j^{(t-1)}$). Since this proposal is symmetric, the acceptance ratio $\alpha_{struct}$ (Algorithm 1, Line 9) simplifies to the ratio of unnormalized posteriors:
\begin{equation}
    \alpha_{struct} = \min\left(1, \frac{P(\mathcal{D} \mid \mathbf{S}(\boldsymbol{\gamma}'), \mathbf{k}(\boldsymbol{\gamma}', \boldsymbol{\theta})) \cdot P(\boldsymbol{\gamma}')}{P(\mathcal{D} \mid \mathbf{S}(\boldsymbol{\gamma}), \mathbf{k}(\boldsymbol{\gamma}, \boldsymbol{\theta})) \cdot P(\boldsymbol{\gamma})} \right)
\end{equation}
where $\mathbf{S}(\boldsymbol{\gamma})$ is the stoichiometry defined by the active subset. Crucially, during a structure move, the auxiliary parameters $\boldsymbol{\theta}$ are held fixed.

\textbf{Parameter Move.} 
We employ a Random Walk Metropolis kernel for the continuous parameters. The proposal distribution is a Gaussian centered at the current value: $\theta'_j \sim \mathcal{N}(\theta_j^{(t-1)}, \sigma_{prop}^2)$. The acceptance ratio is:
\begin{equation}
    \alpha_{param} = \min\left(1, \frac{P(\mathcal{D} \mid \boldsymbol{\gamma}, \boldsymbol{\theta}') P(\boldsymbol{\theta}' \mid \boldsymbol{\gamma})}{P(\mathcal{D} \mid \boldsymbol{\gamma}, \boldsymbol{\theta}) P(\boldsymbol{\theta} \mid \boldsymbol{\gamma})} \right)
\end{equation}
If thermodynamic constraints are violated by $\boldsymbol{\theta}'$, the prior term $P(\boldsymbol{\theta}' \mid \boldsymbol{\gamma})$ evaluates to zero, leading to automatic rejection.

\begin{algorithm} 
    \caption{Bayesian Inference of Reaction Network Topology and Kinetic Parameters}
    \label{alg:pcmcmc}
    \begin{algorithmic}[1] 
        \REQUIRE Data $\mathcal{D}$, Atomic Matrix $\mathbf{A}$, Candidate Reactions $\mathcal{R}$
        \ENSURE Posterior samples of Topology $\mathbf{S}$ and Parameters $\mathbf{k}$
        
        \STATE \textbf{Initialize} $\gamma^{(0)}, \mathbf{k}^{(0)}$ satisfying constraints
        
        \FOR{$t = 1$ to $T$}
            \STATE \textit{// Structure Move (Topology Sampling)}
            \STATE Propose new topology $\gamma'$ by flipping bit $j$ of $\gamma^{(t-1)}$
            \STATE Construct candidate stoichiometry $\mathbf{S}'$ based on $\gamma'$
            
            \IF{$\mathbf{A} \cdot \mathbf{S}'_{\cdot j} \neq \mathbf{0}$}
                \STATE Reject $\gamma'$ (Violation of Mass/Electron Conservation)
            \ELSE
                \STATE Compute Acceptance Ratio $\alpha_{struct}$ with Sparsity Prior
                \STATE Accept $\gamma^{(t)} = \gamma'$ with probability $\min(1, \alpha_{struct})$
            \ENDIF
            
            \STATE
            \STATE \textit{// Parameter Move (Kinetic Constant Sampling)}
            \STATE Propose $\mathbf{k}'$ from proposal distribution $q(\mathbf{k}' \mid \mathbf{k}^{(t-1)})$
            
            \IF{Thermodynamic Loop Constraints ($\Delta G$) violated}
                \STATE Reject $\mathbf{k}'$ (Enforce Thermodynamic Consistency)
            \ELSE
                \STATE Calculate Likelihood $P(\mathcal{D} \mid \mathbf{S}^{(t)}, \mathbf{k}')$ via stiff ODE integration
                \STATE Accept $\mathbf{k}^{(t)} = \mathbf{k}'$ based on Metropolis criterion
            \ENDIF
        \ENDFOR
        
        \STATE \textbf{return} MCMC chains for $\mathbf{S}$ and $\mathbf{k}$
    \end{algorithmic}
\end{algorithm}

\begin{algorithm}[H]
    \caption{CIGP-driven Active Learning}
    \label{alg:cigp}
    \begin{algorithmic}[1]
        \REQUIRE Initial data $\mathcal{D}_{init}$, Physical ODE model $f_{phys}$, Acquisition function $\alpha(\cdot)$, Budget $N_{BO}$
        \ENSURE Optimized process conditions $\mathbf{u}^*$, Calibrated parameters $\mathbf{W}^*$
        
        \STATE $\mathcal{D} \leftarrow \mathcal{D}_{init}$
        
        \FOR{$i = 1$ to $N_{BO}$}
            \STATE \textit{// Hybrid Model Construction}
            \STATE Define Mean Function: $\mu(\mathbf{u}) = f_{phys}(\mathbf{u}; \mathbf{W})$
            \STATE Define Kernel: $k(\mathbf{u}, \mathbf{u}') = \text{RBF}(\mathbf{u}, \mathbf{u}'; \Theta_{gp})$
            
            \STATE
            \STATE \textit{// Joint Training (Constrained Optimization)}
            \STATE Define Objective: $J(\mathbf{W}, \Theta_{gp}) = \mathcal{L}_{GP} - \lambda \sigma_f^2$
            \STATE Solve: $\mathbf{W}^*, \Theta_{gp}^* \leftarrow \text{argmax } J$ 
            \STATE \quad \textbf{subject to} $\sigma_{min}^2 \le \sigma_f^2 \le \epsilon_{tol}$
            
            \STATE
            \STATE \textit{// Physics-Aware Acquisition}
            \STATE Select next experiment point $\mathbf{u}_{new}$:
            \STATE $\mathbf{u}_{new} = \text{argmax}_{\mathbf{u} \in \mathcal{U}} \alpha(\mathbf{u} \mid \mathcal{D}, \mathbf{W}^*)$ \textit{(e.g., GWU, PC-EI)}
            
            \STATE
            \STATE \textit{// Experimental Feedback}
            \STATE Conduct experiment: $y_{new} \leftarrow \text{LabOracle}(\mathbf{u}_{new})$
            \STATE Update dataset: $\mathcal{D} \leftarrow \mathcal{D} \cup \{\mathbf{u}_{new}, y_{new}\}$
        \ENDFOR
        
        \STATE \textbf{return} Best found condition $\mathbf{u}^*$ and Parameters $\mathbf{W}^*$
    \end{algorithmic}
\end{algorithm}

\textbf{Integrated Workflow.} The proposed framework operates in two coupled phases to achieve robust autonomous discovery. 
First, \textbf{PC-MCMC (Algorithm 1)} is employed to identify the most probable reaction topology and estimate initial kinetic parameters from preliminary data. 
Second, \textbf{CIGP-driven Active Learning (Algorithm 2)} utilizes the identified topology (embedded in $f_{phys}$) to construct a hybrid surrogate model. This model guides the experimental design to simultaneously refine the kinetic parameters $\mathbf{W}$ and optimize process conditions $\mathbf{u}$, as detailed below.

\begin{figure}[h]
    \centering
    \includegraphics[width=1.0\textwidth]{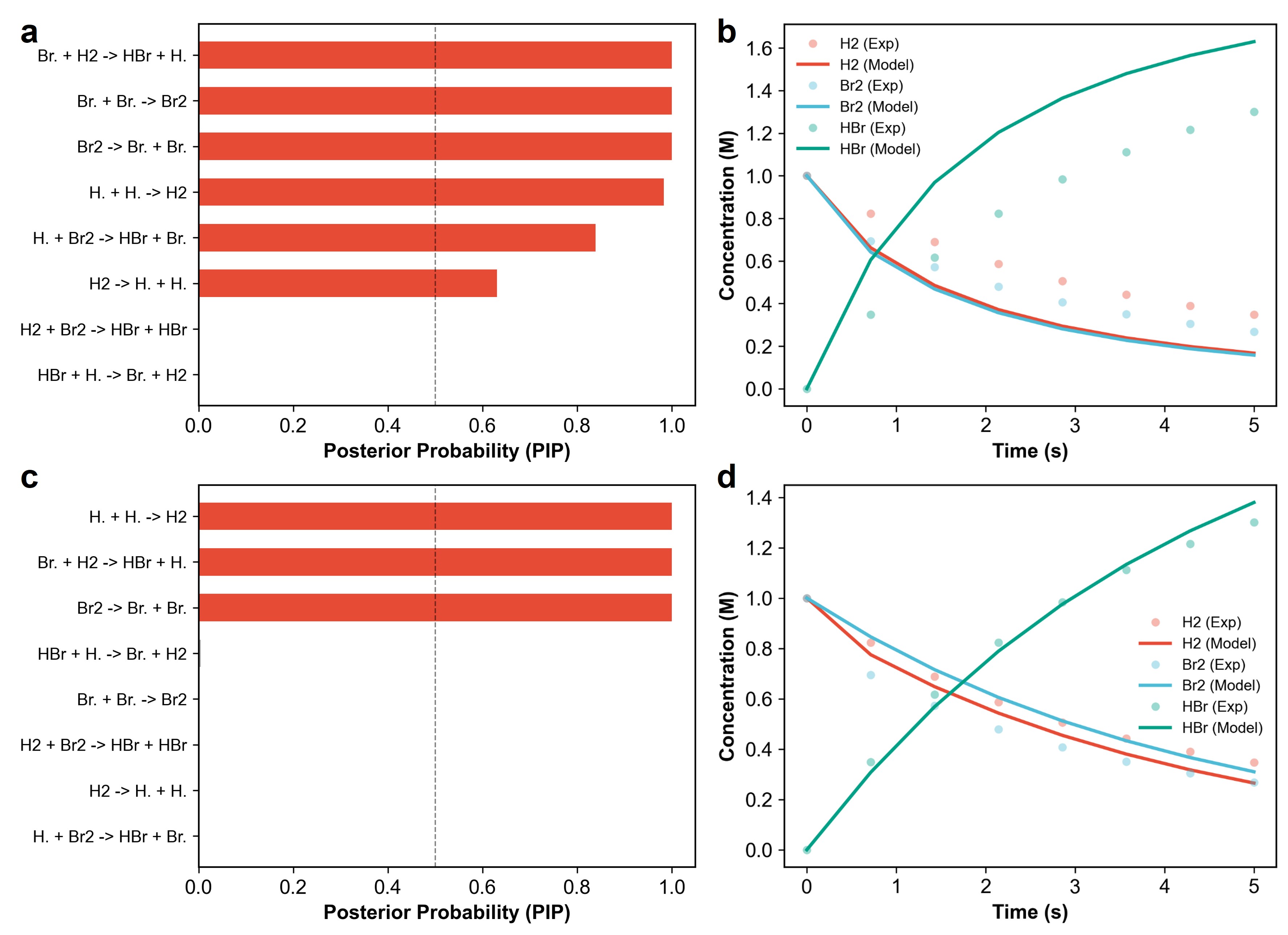}
    
    \caption{\textbf{Visualization of structural failure modes under ablation settings.} Panels a and b depict the absence of the sparsity prior where the sampler converges to a dense topology in panel a and assigns high posterior probability to kinetically prohibited pathways to minimize residual error in panel b. Panels c and d illustrate the absence of thermodynamic constraints where the principle of microscopic reversibility is violated in panel c as forward chain initiation is selected with a PIP near 1.0 while the reverse termination is suppressed with a PIP near 0.0 resulting in a physically inconsistent mechanism despite accurate trajectory fitting in panel d.}
    \label{fig:ablation_vis}
\end{figure}

\textbf{Visual Analysis of Failure Modes.} The structural consequences of relaxing inductive biases are further elucidated through posterior landscape visualization as shown in Figure \ref{fig:ablation_vis}. In the absence of the sparsity-inducing prior as illustrated in panels a and b of Figure \ref{fig:ablation_vis} a distinctively dense topology is retrieved. High posterior inclusion probabilities exceeding 0.6 are assigned indiscriminately to candidate reactions including kinetically prohibitive pathways such as the direct dissociation of hydrogen. While reasonable alignment with observational data is maintained in the reconstructed trajectory this is attributed to overfitting wherein spurious structural complexity is recruited to absorb measurement noise rather than isolating the causal signal. Conversely the violation of microscopic reversibility upon the removal of thermodynamic constraints is starkly illustrated in panels c and d of Figure \ref{fig:ablation_vis}. A structural asymmetry is observed where forward reactions such as chain initiation are identified with unit probability whereas the corresponding reverse steps are completely suppressed with a posterior inclusion probability approaching zero effectively decoupling the equilibrium relationship. Although a tight numerical fit is achieved via parameter drift the loss of detailed balance renders the discovered mechanism physically inconsistent and unreliable for extrapolation beyond the training regime.

\paragraph{Implementation Details.}
For the mechanism discovery baselines, \textbf{SINDy} was configured with a polynomial feature library (degree $\le 3$) and the STLSQ optimizer (threshold $\lambda=0.1$). \textbf{PySR} was trained for 100 iterations using default binary operators $(+, -, *, /)$ and a complexity penalty of 0.001. For the $H_2+Br_2$ benchmark, initial concentrations were sampled via LHS from $[0.1, 5.0]$ M and trajectories were generated by integrating the mass-action ODE. The \textbf{Styrene Epoxidation Oracle} simulates a competitive reaction network: main epoxidation ($k_1=10^6 e^{-55000/RT}$) and side hydrolysis ($k_2=10^{10} e^{-85000/RT}$), subject to standard Arrhenius kinetics. The released code includes scripts for Fig.~3, Fig.~4, the acquisition comparison, and smoke-test verification; generated CSV summaries are written under \texttt{examples/outputs/}.

\section{Computational Complexity and Runtime Analysis}
\label{app:computational_cost}

To assess the practical feasibility of the proposed framework, we evaluated the wall-clock execution time of both the PC-MCMC (Structure Discovery) and CIGP (Active Learning) modules. All experiments were conducted on a standard personal laptop equipped with an Intel Core i7-12700H CPU (2.30 GHz) and 32 GB RAM, without utilizing any GPU acceleration.

\subsection{Runtime Breakdown}

Table~\ref{tab:runtime} summarizes the computational cost compared to standard baselines.

\begin{table}[h]
    \centering
    \caption{Average wall-clock time for model inference and optimization tasks.}
    \label{tab:runtime}
    \begin{tabular}{lccc}
    \toprule
    \textbf{Task / Method} & \textbf{Computational Load} & \textbf{Runtime} & \textbf{Hardware} \\
    \midrule
    \multicolumn{4}{l}{\textit{Phase 1: Mechanism Discovery ($H_2+Br_2$ Benchmark)}} \\
    PC-MCMC (Ours) & 10,000 MCMC steps (w/ Stiff ODE) & $28.5 \pm 2.1$ min & CPU \\
    SINDy (Baseline) & Sparse Regression (STLSQ) & $< 1.0$ sec & CPU \\
    PySR (Baseline) & Symbolic Regression (100 gens) & $5.2 \pm 0.5$ min & CPU \\
    \midrule
    \multicolumn{4}{l}{\textit{Phase 2: Closed-Loop Optimization (Styrene Epoxidation)}} \\
    CIGP (Ours) & Hyperparam Opt + Acquisition (per iter) & $1.2 \pm 0.3$ sec & CPU \\
    Standard GP (Baseline) & Analytical Inference (per iter) & $< 0.1$ sec & CPU \\
    \bottomrule
    \end{tabular}
\end{table}

\subsection{Analysis of Computational Overhead}

\textbf{Structure Discovery (PC-MCMC).} 
The PC-MCMC sampler requires solving a stiff ODE system at each Metropolis-Hastings step to evaluate the likelihood. While this incurs a higher computational cost ($\approx 30$ minutes) compared to regression-based methods like SINDy, it is negligible in the context of mechanistic discovery, which is typically a one-time offline process. Furthermore, the fixed-dimensional parameter space of our Spike-and-Slab formulation avoids the complexities of trans-dimensional sampling (e.g., RJMCMC), ensuring convergence within a reasonable timeframe on personal computing devices.

\textbf{Active Learning (CIGP).} 
For the online optimization phase, the CIGP inference entails numerical integration of the mean function $f_{phys}$. Although this increases the cost per iteration to $\approx 1.2$ seconds (compared to $<0.1$ seconds for a standard GP), this overhead is small relative to laboratory reaction and analysis times. The full PC-MCMC-CIGP loop should nevertheless be viewed as a two-stage workflow: PC-MCMC is an offline structure-discovery step, while CIGP acquisition is the online step used between experiments.

\end{document}